\documentclass{article}
\usepackage{multirow}
\usepackage{ijcai26}

\usepackage{times}
\usepackage{soul}
\usepackage{url}
\usepackage[hidelinks]{hyperref}
\usepackage[utf8]{inputenc}
\usepackage[small]{caption}
\usepackage{graphicx}
\usepackage{amsmath}
\usepackage{amsthm}
\usepackage{booktabs}
\usepackage{algorithm}
\usepackage{algorithmic}
\usepackage{amssymb}
\usepackage[switch]{lineno}
\usepackage{colortbl}   
\usepackage{booktabs}
\usepackage[table]{xcolor}
\usepackage{comment}

\title{UniRoute: Unified Routing Mixture-of-Experts for Modality-Adaptive Remote Sensing Change Detection}

\author{
	Qingling Shu$^{1}$\thanks{sql@stu.ahu.edu.cn.} \and
	Sibao Chen$^{1}$\thanks{Corresponding author.} \and
	Wei Lu$^{1}$ \and
	Zhihui You$^{2}$ \and
	Chengzhuang Liu$^{1}$\\[2pt]
	{\affiliations
		$^{1}$MOE Key Lab of ICSP, IMIS Lab of Anhui Province, Anhui Provincial Key Lab of Multimodal Cognitive Computation,School of Computer Science and Technology, Anhui University, Hefei, China\\
		$^{2}$School of Public Safety and Emergency Management, Anhui University of Science and Technology, Hefei, China\\
	}
}

\begin{document}

\maketitle

\begin{abstract}
Current remote sensing change detection (CD) methods mainly rely on specialized models, which limits the scalability toward modality-adaptive Earth observation.
For homogeneous CD, precise boundary delineation relies on fine-grained spatial cues and local pixel interactions,
whereas heterogeneous CD instead requires broader contextual information to suppress speckle noise and geometric distortions.
Moreover, difference operator (e.g. Subtraction) works well for aligned homogeneous images but introduces  artifacts in cross-modal or geometrically misaligned scenarios.
Across different modality  settings, specialized models based on static backbones or fixed difference operations often prove insufficient.
To address this challenge, we propose UniRoute, a unified framework for modality-adaptive learning by reformulating feature extraction and fusion as conditional routing problems. We introduce an Adaptive Receptive Field Routing MoE (AR²-MoE) module to  disentangle local spatial details from global semantic context, and a Modality-Aware Difference Routing MoE (MDR-MoE) module to adaptively select the most suitable fusion primitive at each pixel.
In addition, we propose a Consistency-Aware Self-Distillation (CASD) strategy that stabilizes unified training under data-scarce heterogeneous settings by enforcing multi-level consistency.
Extensive experiments on five public datasets demonstrate that UniRoute achieves strong overall performance,
with a favorable accuracy--efficiency trade-off under a unified deployment setting.



\end{abstract}

\section{Introduction}
\begin{figure*}[t]
	\centering
	\includegraphics[width=1\linewidth]{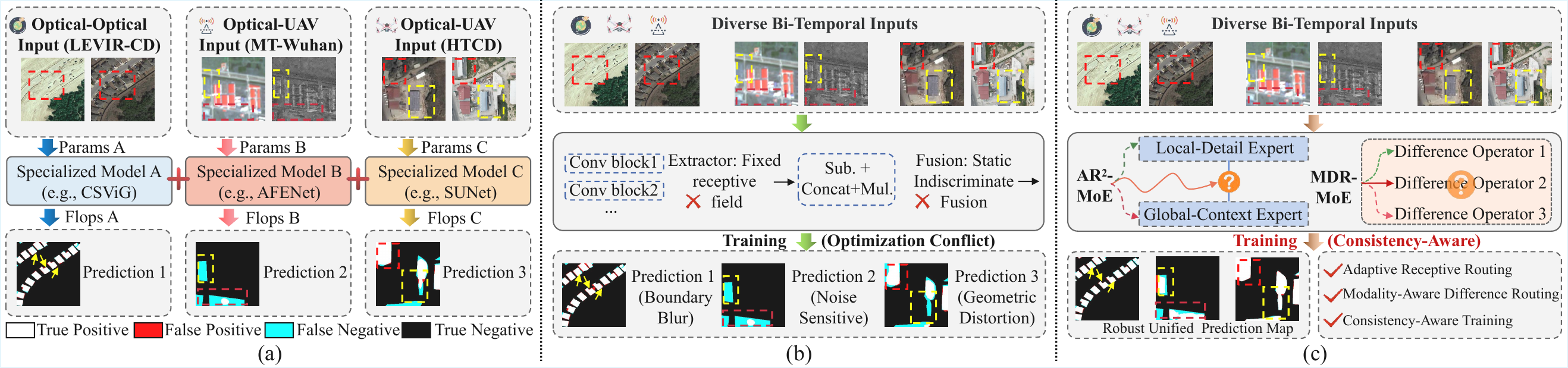} 
	\vspace{-0.6cm}
	\caption{(a) Isolated specialized methods for different modality pairs. (b) Static unified baselines suffering from fixed receptive fields and indiscriminate fusion. (c) Our UniRoute framework with modality-adaptive routing and consistency-aware training.}
	\vspace{-0.3cm}
	\label{fig:Fig1}
\end{figure*}
Change detection (CD) is a fundamental task in Earth observation~\cite{Into1}. 
With the proliferation of multi-source sensors, real-world applications increasingly demand the processing of diverse data modalities, ranging from homogeneous optical image pairs to heterogeneous combinations like Optical-synthetic aperture radar (SAR) and Optical-Unmanned Aerial Vehicle (UAV).
Traditionally, CD models are designed in a  specialized manner. 
For homogeneous CD, siamese architectures with shared encoders and explicit difference operations are commonly adopted to model intensity changes~\cite{SNUNet,DMINet,ELGC-Net,SFEARNet}. 
In contrast, heterogeneous CD cannot rely on direct comparison due to pronounced domain gaps. 
To handle this issue, many approaches use pseudo-siamese fusion networks. 
These networks learn non-linear semantic mappings or perform style transfer across modalities~\cite{BAN,AFENet,XiongAn,HRSICD}.
Although these designs perform well in specific settings (Fig.~\ref{fig:Fig1}(a)),
they typically require separate models for different modality pairs in a unified setting,
resulting in increased maintenance and training costs as more modalities are considered.

As shown in Fig.~\ref{fig:Fig1}(b), unified training exposes  challenges that arise in practice.
One issue appears during feature extraction and is closely related to receptive field design,  which  determines the balance between local spatial details and global contextual information. Homogeneous inputs benefit from fine-grained spatial details, which are important for delineating building boundaries.
Heterogeneous data present a different situation. For example, Optical–SAR pairs are frequently affected by speckle noise, where broader semantic context becomes more useful than local details. 
Methods such as A2Net~\cite{A2Net} and PKINet~\cite{PKINet} combine features produced by convolutional kernels of different sizes.
Other architectures, including  CDMamba~\cite{CDMamba} and BSSMamba~\cite{BSSMamba} rely on mechanisms such as modulation or attention to mix local and global information. Although these designs differ in implementation, they all follow a feature integration strategy, which requires the network to accommodate signals with competing characteristics.  In SAR data, high-frequency noise captured by local branches can therefore interfere with global semantic representations.
Moreover,
the behavior of difference operations varies noticeably across modalities.
In homogeneous CD, simple pixel-wise interactions often work well, and subtraction can highlight boundary changes.
This behavior does not generalize to heterogeneous pairs.
For Optical–SAR data, subtraction tends to amplify speckle noise.
Geometric misalignment also leads to false responses near object boundaries in Optical-UAV pairs.
Some recent methods attempt to reduce this sensitivity by combining multiple operations.
Models include CSViG~\cite{CSViG} and ChangeCLIP~\cite{ChangeCLIP} aggregate subtraction, concatenation, and related operators within a unified pipeline. This indiscriminate aggregation  leads the noisy branches to influence the feature representation when applied  in heterogeneous settings.

To address the structural challenges, we argue that a unified model should avoid static compromises. The model should instead act as a dynamic system that can reconfigure itself. Fig.~\ref{fig:Fig1}(c) illustrates this idea. Inspired by Mixture-of-Experts (MoE)~\cite{MoE1}, we propose UniRoute, a modality-adaptive MoE framework built on dynamic routing.
We first address the receptive field conflict. We replace standard convolutional blocks with the Adaptive Receptive Field Routing MoE ($\text{AR}^2$-MoE). Unlike image-level routing approaches~\cite{MoE2,MoE3}, $\text{AR}^2$-MoE applies hard routing at the pixel level. This design separates local spatial details, which are critical for boundary detection, from global semantic context, which supports noise suppression.
We next consider the operation conflict. We introduce the Modality-Aware Difference Routing MoE (MDR-MoE). MDR-MoE does not combine all fusion primitives by default. Instead, it selects fusion branches based on the input modality. This process removes operations that are likely to introduce noise during feature fusion.
Training a unified model across modalities also raises overfitting concerns. This issue is especially severe in data-scarce heterogeneous settings. To mitigate this risk, we propose a Consistency-Aware Self-Distillation (CASD) paradigm. CASD uses self-distillation to enforce geometric invariance. It also introduces decision-level and statistical consistency to stabilize training.
Our  contributions are summarized as follows:

\begin{itemize}
	\item We identify the intrinsic conflicts in receptive fields and difference operations for unified CD, and propose UniRoute, a modality-adaptive framework that resolves these conflicts through dynamic expert routing.
	\item We design $\text{AR}^2$-MoE and MDR-MoE, which introduce pixel-wise hard routing to adaptively select optimal spatial contexts and fusion primitives, respectively.
	\item We propose the CASD training paradigm, which mitigates the small-sample overfitting problem by internalizing multi-level consistency via self-distillation.
	\item 
	We further demonstrate favorable deployment efficiency: UniRoute supports real-time inference on a single GPU
	and avoids the system overhead of maintaining or swapping multiple specialized models.
	

\end{itemize}

\section{Related Work}
\subsection{Remote Sensing Change Detection}
Existing CD methods can be broadly categorized into homogeneous and heterogeneous approaches based on the input modalities.
For homogeneous data, siamese networks are the mainstream choice.  Methods like  TFI-GR~\cite{TFI-GR}, PGPANet~\cite{PGPANet} and DMINet~\cite{DMINet} employ robust backbone networks to extract features and utilize algebraic subtraction to highlight changes.  
Recent Transformers~\cite{Changeformer,BIT,ELGC-Net} and State Space Models ChangeMamba~\cite{ChangeMamba} improve long-range context modeling,
while SCD-SAM~\cite{SCD-SAM} utilizes the pre-trained knowledge of the Segment Anything Model for better semantic alignment. 
Foundation models like Changen2~\cite{Changen2} further leverage large-scale pre-training for universal representations.
For heterogeneous data, the domain gap renders simple subtraction ineffective. Methods like SUNet~\cite{SUNet}, ChangeLN~\cite{ChangeLN}, AFENet~\cite{AFENet} and  HeteCD~\cite{XiongAn} typically rely on style transfer, feature mapping, or metric learning to align multimodal features.
However, these divide-and-conquer methods are unscalable. Homogeneous models fail on SAR data due to noise sensitivity, while heterogeneous models often lack boundary precision on optical imagery. Our UniRoute unifies these scenarios into a single framework, dynamically selecting the optimal difference primitive via the MDR-MoE.
\subsection{Mixture-of-Experts in  Computer Vision}
MoE scales model capacity without increasing inference cost by conditionally activating a subset of experts~\cite{MoE1}. In computer vision, SCNet~\cite{DeepMoE} reforms standard convolutions via a self-calibration operation that  expands the field-of-view. 
Recent vision MoEs (e.g., AdaMV-MoE~\cite{MoE2}, MLoRE~\cite{MoE3}) have explored image-level or patch-level routing.  DAMEX~\cite{DAMEX} introduces dataset-aware routing for universal object detection. However, most approaches employ soft gating or shared expert architectures (e.g., identical convolution blocks), which leads to feature blurring and insufficient specialization. This is critical for dense prediction tasks like CD, where precise boundary delineation is paramount.
In contrast, our AR$^2$-MoE introduces a pixel-wise hard routing mechanism via the Straight-Through Estimator (STE)~\cite{Gumbel_Softmax}. 
\begin{figure*}[t]
	\centering
	\includegraphics[width=1\linewidth]{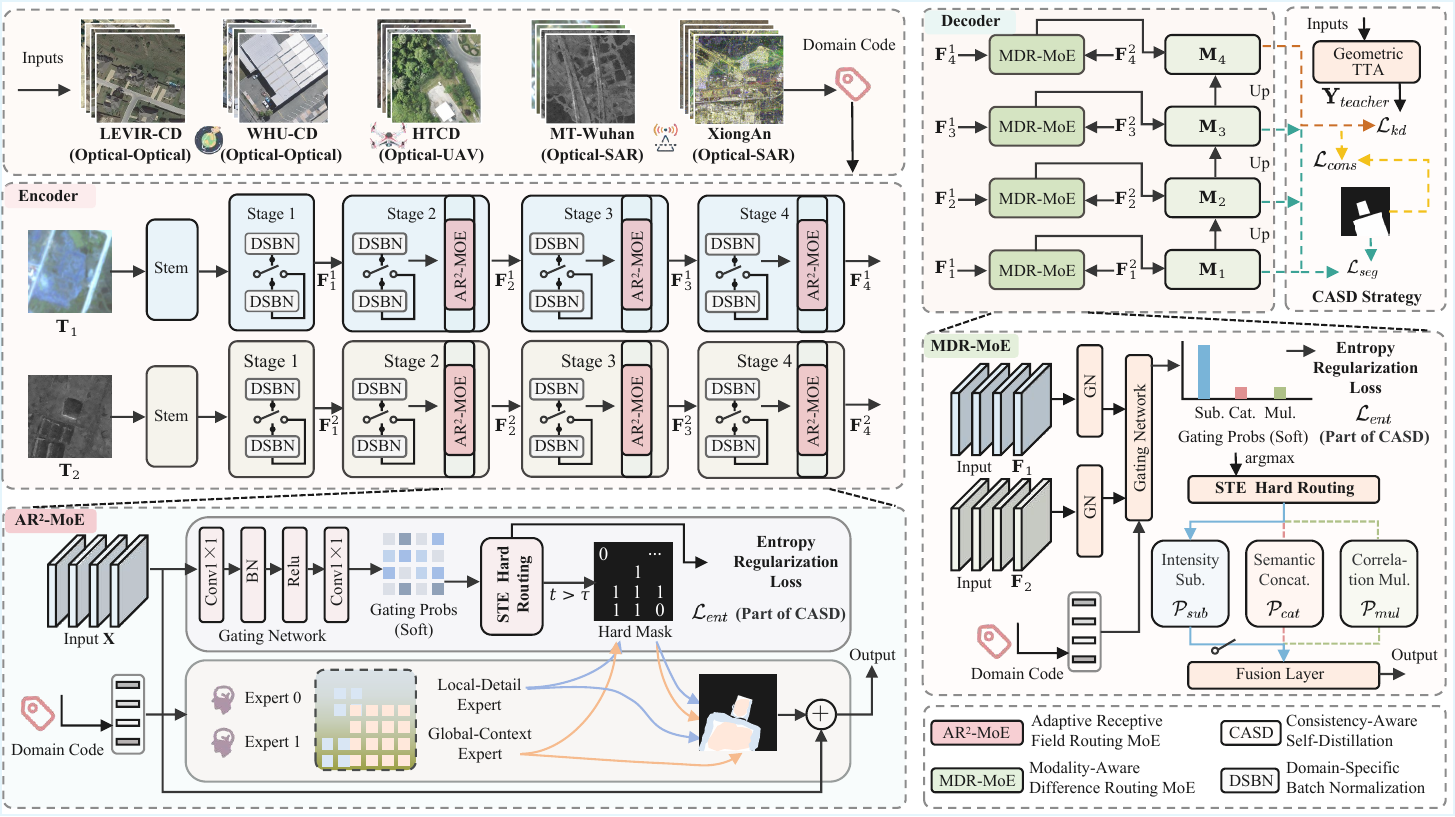} 
	\vspace{-0.7cm}
	\caption{
		Overview of UniRoute. It features an $\text{AR}^2$-MoE for adaptive receptive field selection in the encoder and an MDR-MoE for optimal fusion primitive routing in the decoder, optimized via the CASD strategy.	
	}
\vspace{-0.2cm}
	\label{fig:overview}
\end{figure*}
Unlike coarse-grained routing that may introduce block artifacts, our method achieves fine-grained adaptation by explicitly disentangling high-frequency local details from global context at every spatial location, resolving the receptive field conflict inherent in multi-modal data.
\subsection{Unified Modeling \& Training Strategies}
Developing a unified model for diverse tasks is a long-standing goal. Pioneering work like  Perceiver IO~\cite{Perceiver_IO} utilizes a shared Transformer backbone to handle multi-modal inputs via tokenization. In remote sensing, SM3Det~\cite{sm3det} explores unified object detection across optical and SAR domains.
A key challenge in unified training is the optimization conflict arising from task heterogeneity.
While multi-task learning methods address this by balancing gradients~\cite{Gradnorm}, and Test-Time Adaptation (TTA)~\cite{Tent} improves robustness during inference, these approaches often require complex optimization or online updates.
Alternatively, we propose a CASD paradigm. Instead of heavy optimization tricks, we leverage self-supervised geometric consistency to regularize training. 

\section{Methodology}

\subsection{Overview}
Given a collection of  remote sensing CD datasets $\mathcal{D} = \{D_1, \dots, D_N\}$, where each sample consists of a bi-temporal image pair $(\mathbf{T}_1, \mathbf{T}_2)$ and a change map $\mathbf{GT}$. The modalities of $\mathbf{T}_1$ and $\mathbf{T}_2$ can be either homogeneous (e.g., Optical-Optical) or heterogeneous (e.g.,Optical-UAV, Optical-SAR).
We propose UniRoute, a unified modality-adaptive framework shown in Fig.~\ref{fig:overview}. It introduces the AR$^2$-MoE to disentangle receptive field requirements and the MDR-MoE to dynamically select optimal fusion primitives.
Furthermore, to mitigate overfitting on small-sample heterogeneous datasets, we introduce a CASD training paradigm, which internalizes geometric invariance and aligns cross-modal semantics.

\subsection{Adaptive Receptive Field Routing MoE}
\label{sec:ar_moe}
To address modality-dependent receptive field conflicts, we introduce AR²-MoE, which treats receptive field selection as an input-conditioned routing problem. Rather than statically aggregating multi-scale features, AR²-MoE performs pixel-wise hard routing between two experts with complementary receptive field properties, allowing each spatial location to adaptively select an appropriate context.

\noindent\textbf{Decoupled Receptive-Field Experts.} 
Given the intermediate feature map $\mathbf{F}_{s}^{1},\mathbf{F}_{s}^{2} \in \mathbb{R}^{B \times C_s \times H_s \times W_s}$ form the temporal branch $s \in \{1, 2, 3, 4\}$, we omit batch and stage indices for brevity and denote the input as $\mathbf{X}$.
The AR$^2$-MoE  aims to learn a dynamic mapping 
by selectively activating experts from a decomposed functional space. 
We construct two experts with complementary receptive field behaviors. The \textit{Local-Detail Expert} focuses on preserving fine-grained spatial structures using a compact convolutional operator, which is effective for boundary-sensitive optical imagery. 
\begin{equation}
E_{lde}(\mathbf{X}) = \Phi_{local}(\mathbf{X}; \mathbf{W}_{3 \times 3}),
\end{equation}
where $\Phi_{local}$ denotes a depth-wise separable convolution block with a small kernel size.
The \textit{Global-Context Expert} uses a dilated, decomposed convolution sequence to emphasize broader contextual information, helping reduce modality-specific noise such as SAR speckle.
Instead of computationally prohibitive self-attention, we approximate global dependencies via a decomposed convolution sequence:
\begin{equation}
E_{gce}(\mathbf{X}) = \mathcal{F}_{pw}(\mathcal{F}_{dilated}(\mathcal{F}_{dw}(\mathbf{X}))) \odot \mathbf{X},
\end{equation}
where $\mathcal{F}_{dw}$ denotes a depth-wise convolution, $\mathcal{F}_{dilated}$ is a depth-wise dilated convolution with dilation rate $d=3$ and $\mathcal{F}_{pw}$ represents a point-wise convolution for channel mixing. The operator $\odot$ represents element-wise modulation. With this decomposition, the expert primarily attends to features from a large receptive field. This reduces the influence of local noise and preserves the global semantic information.

\noindent\textbf{Entropy-Constrained Hard Routing.}
To dynamically assign these experts, we employ a grid-level gating network $G$ conditioned on  local visual features $\mathbf{X}$: 
\begin{equation}
\mathbf{g}
= \sigma\!\left(
\mathcal{W}_{g}
\big(
\gamma(\mathbf{z}) \odot \phi(\mathbf{X})
+ \beta(\mathbf{z})
\big)
\right).
\end{equation}
Here, $\phi(\cdot)$ denotes a lightweight feature projection implemented as a $1\times1$ convolution, and $\gamma(\mathbf{z})$ and $\beta(\mathbf{z})$ perform channel-wise modulation on the projected features, while the gating weights $W_g$ map the modulated features to expert selection logits.
$\sigma$ is the Sigmoid function 
and $\mathbf{z}$ is a domain code representing the source domain, which is available as domain metadata and used as a conditioning prior.
We find that soft gating can mix the responses of different experts, making their roles less distinct. To preserve expert specialization, we use discrete hard routing based on the Straight-Through Estimator (STE). The binary routing mask $\mathbf{M}$ is computed as:
\begin{equation}
    \mathbf{M} = \mathbb{I}(\mathbf{g} > 0.5) - \text{detach}(\mathbf{g}) + \mathbf{g},
\end{equation}
where $\mathbb{I}(\cdot)$ is the indicator function. The term $\text{detach}(\mathbf{g})$ stops the gradient flow, allowing $\mathbf{M}$ to behave as a binary mask during the forward pass while enabling gradients to flow through the continuous probability $\mathbf{g}$ during backward propagation.
We adopt STE-based hard routing to enable discrete expert selection while maintaining end-to-end differentiability. During training, routing decisions are binary in the forward pass, while gradients are propagated through the continuous gating probabilities.
The final output is dynamically assembled: 
\begin{equation}
    \mathbf{y} = (1 - \mathbf{M}) \odot E_{lde}(\mathbf{X}) + \mathbf{M} \odot E_{gce}(\mathbf{X}) + \mathbf{X}.
\end{equation}
This  design ensures that in ambiguous regions where experts might provide conflicting signals, the network preserves the original feature identity $\mathbf{X}$, stabilizing the gradient flow.


\subsection{Modality-Aware Difference Routing MoE}
\label{sec:poly_diff}

Difference modeling in unified CD is highly modality-dependent. Static fusion of multiple operators often leads to incompatible responses across heterogeneous inputs. To address this issue, we propose MDR-MoE, which formulates difference modeling as a conditional routing problem.

\noindent\textbf{Library of Differentiable Primitives.} 
Given bi-temporal features $\mathbf{F}_s^1, \mathbf{F}_s^2$ at stage $s$, we  simplify the notation to $\mathbf{F}_1, \mathbf{F}_2$.
To enable modality-dependent selection rather than indiscriminate fusion, we construct a compact library of differentiable primitives, each encoding a distinct inductive bias for change modeling. Formally, we denote the primitive set as $\mathcal{P}=\{\mathcal{P}_k\}_{k=1}^K$:
including subtraction, concatenation, and multiplicative interaction. Each primitive captures a distinct inductive bias for change modeling, and a lightweight projection layer is applied for feature alignment.

\noindent\textbf{Modality-Aware Hard Routing.}
Instead of indiscriminate fusion, which cannot resolve operation conflicts across modalities, MDR-MoE employs a discrete routing mechanism.
We utilize a gating network $G$ to predict the selection probability $\mathbf{\Pi} \in \mathbb{R}^{B \times K \times H \times W}$ based on the input features and a domain code.
\begin{equation}
\mathbf{\Pi}
= \sigma\!\left(
\mathcal{W}_{g}
\Big(
\gamma(\mathbf{z}) \odot \phi([\mathbf{F}_1, \mathbf{F}_2])
+ \beta(\mathbf{z})
\Big)
\right),
\end{equation}
Here, $\mathbf{z}$ is shared with AR$^2$-MoE as a domain conditioning code to ensure consistent operator selection.
To ensure strict operation decoupling (i.e., actively pruning noisy branches), we apply Top-1 Hard Routing via the STE. The binary selection mask $\mathbf{Z} \in \{0, 1\}^{B \times K \times H \times W}$ is:
\begin{equation}
    \mathbf{Z} = \text{OneHot}(\arg\max_k({\mathbf{\Pi}})) - \text{detach}(\mathbf{\mathbf{\Pi}}) + \mathbf{\mathbf{\Pi}}.
\end{equation}
The final difference feature $\mathbf{M}_{diff}$ is dynamically assembled by selecting the optimal primitive for each spatial location:
\begin{equation}
    \mathbf{M}_{diff} = \sum_{k=1}^{K} \mathbf{Z}_k \odot \mathcal{P}_k(\mathbf{F}_1, \mathbf{F}_2),
\end{equation}
where $\mathbf{Z}_k$ is the one-hot hard mask for the $k$-th primitive (corresponding to $\mathcal{P}_{sub}, \mathcal{P}_{cat}, \mathcal{P}_{mul}$, respectively). 
With the proposed routing, subtraction is used for aligned homogeneous inputs, while fusion-based primitives are applied to heterogeneous or geometrically misaligned data in the same model.

\subsection{Consistency-Aware Self-Distillation (CASD)}
\label{sec:casd}

Training a unified model on heterogeneous datasets is challenging due to inconsistent supervision across modalities under limited data, which leads to unstable optimization and unreliable expert routing. We introduce CASD as a training regularization strategy to stabilize optimization by enforcing multi-level consistency.

\noindent\textbf{Geometric Consistency via Self-Distillation.}
Standard supervision is sensitive to geometric variations and easily overfits in data-scarce settings. We therefore enforce prediction consistency under spatial transformations through self-distillation with Test-Time Augmentation (TTA). Specifically, given a transformation $\mathcal{T}$ (e.g., horizontal flip), the model aggregates predictions from the original and transformed views to construct a robust soft target:
\begin{equation}
\mathbf{Y}_{teacher} = \frac{1}{2}(\sigma(\mathcal{M}(\mathbf{T}_1,\mathbf{T}_2)) + \mathcal{T}^{-1}\!(\sigma(\mathcal{M}(\mathcal{T}(\mathbf{T}_1,\mathbf{T}_2))))).
\end{equation}
Here, $\mathcal{M}$ denotes the model and $\mathcal{T}^{-1}$ represents the inverse spatial transformation to align the predictions. The student model (standard forward pass) is then optimized via MSE loss to match this soft target: 
\begin{equation}
\mathcal{L}_{kd} = \|\sigma(\mathcal{M}(\mathbf{X})) - \mathbf{Y}_{teacher}\|^2,
\end{equation}
which internalizes geometric invariance into the model weights and mitigates overfitting on small-sample datasets.

\begin{table*}[t]
	\scalebox{0.77}{
		\centering
		\begin{tabular}{lcccccccccccccc}
			\toprule
			\multirow{2}{*}{Model} &\multirow{2}{*}{Setting}& Params & FLOPs & \multicolumn{2}{c}{LEVIR-CD} & \multicolumn{2}{c}{WHU-CD} & \multicolumn{2}{c}{HTCD} & \multicolumn{2}{c}{MT-Wuhan} & \multicolumn{2}{c}{XiongAn} & \multirow{2}{*}{Avg F1 $\uparrow$} \\
			\cmidrule(lr){3-4} \cmidrule(lr){5-6} \cmidrule(lr){7-8} \cmidrule(lr){9-10} \cmidrule(lr){11-12} \cmidrule(lr){13-14} &
			& (M) $\downarrow$   & (G) $\downarrow$  & F1 $\uparrow$           & IoU $\uparrow$         & F1 $\uparrow$        & IoU $\uparrow$       & F1 $\uparrow$         & IoU $\uparrow$        & F1   $\uparrow$        & IoU $\uparrow$        & F1  $\uparrow$         & IoU  $\uparrow$        &                         \\
			\midrule
			
			\multicolumn{14}{l}{\textit{Single-Modal Specialists (Trained Individually)}} \\
			\hline

			\rowcolor{gray!20} \multicolumn{15}{l}{BCD}    \\ 
			
			CSViG {\scriptsize(TGRS 23)}    & Indiv.     & 38.0   & 203.0 & 91.52 & 84.37 & 90.57 & 82.76 & 91.16 & 83.76 & 54.80 & 37.74 & 75.17 & 60.22 & 80.64 \\
			ChangeMamba {\scriptsize(TGRS 24)}& Indiv.   & 84.7   & 179.3 & 91.18 & 83.80 & 94.19 & 89.02 & 92.54 & 86.12 & 56.44 & 39.31 & 78.63 & 64.78 & 82.60 \\
			LGCANet {\scriptsize(TGRS 25)}   & Indiv.     & 31.0   & 56.6  & 89.96 & 81.75 & 94.23 & 89.09 & 90.90 & 83.52 & 52.21 & 35.33 & 78.95 & 65.22 & 81.25 \\
			LWGANet {\scriptsize(AAAI 26)}   & Indiv.     & 16.1   & 22.1  & \textbf{92.42} & \textbf{85.90} & \underline{95.24} & \underline{90.92} & 92.21 & 85.55 & 55.78 & 38.68 & 79.48 & 65.94 & 83.03 \\
			
			\hline
			\rowcolor{gray!20} \multicolumn{15}{l}{MCD}    \\
			AFENet {\scriptsize(TGRS 24)}  & Indiv.       & 39.7   & 99.6  & 89.61 & 81.18 & 92.39 & 85.86 & \underline{95.73} & \underline{91.82} & \underline{59.82} & \underline{42.67} & 80.20 & 66.95 & \underline{83.55} \\
			Bi-DiffCD {\scriptsize(IJCAI 25)}  & Indiv.   & 68.3   & 76.5  & 89.86 & 81.58 & 93.49 & 87.78 & 92.54 & 86.12 & 52.60 & 35.69 & 79.13 & 65.47 & 81.52 \\
			HRSICD {\scriptsize(ISPRS 25)}   & Indiv.     & 20.5   & 89.3  & 89.42 & 80.87 & 92.86 & 86.67 & 93.11 & 87.10 & 54.82 & 37.76 & \underline{80.60} & \underline{67.50} & 82.16 \\ 
			HeteCD {\scriptsize(ISPRS 25)}  & Indiv.      & 50.2   & 58.2  & 90.20 & 82.19 & 93.36 & 87.55 & 94.34 & 89.30 & 56.38 & 39.25 & 80.34 & 67.14 & 82.92 \\
			\textbf{Ensemble of Best Specialists} & Indiv. & 132.1 & 332.7 & 92.42 & 85.90 & 95.24 & 90.92 & 95.73 & 91.82 & 59.82 & 42.67 & 80.60 & 67.50 & 84.76 \\ \hline
			
			\multicolumn{13}{l}{\textit{Unified Baselines (Re-implemented)}} \\
			\rowcolor{gray!20} \multicolumn{15}{l}{BCD}    \\
			CSViG {\scriptsize(TGRS 23)} \dag   & Unified     & 31.0   & 203.0 & 89.84 & 81.56 & 88.47 & 79.33 & 85.01 & 73.93 & 51.54 & 34.71 & 78.18 & 64.17 & 78.61 \\
			ChangeMamba {\scriptsize(TGRS 24)}\dag & Unified  & 84.7   & 179.3 & 89.35 & 80.75 & 92.88 & 86.70 & 87.43 & 77.66 & 53.88 & 36.87 & 76.43 & 61.85 & 79.99 \\
			LGCANet {\scriptsize(TGRS 25)}\dag  & Unified      & 31.0   & 56.6  & 88.42 & 79.25 & 91.00 & 83.48 & 84.75 & 73.53 & 49.68 & 33.05 & 75.49 & 60.63 & 77.87 \\
			LWGANet {\scriptsize(AAAI 26)} \dag   & Unified    & 16.1   & 22.1  & 90.32 & 82.35 & 93.48 & 87.70 & 86.84 & 76.75 & 52.38 & 35.48 & 76.77 & 62.29 & 79.96 \\
			
			\hline
			\rowcolor{gray!20} \multicolumn{15}{l}{MCD}    \\
			AFENet {\scriptsize(TGRS 24)} \dag   & Unified     & 39.7   & 99.6  & 88.49 & 79.36 & 90.36 & 82.42 & 88.62 & 79.56 & 56.11 & 39.19 & 78.71 & 64.89 & 80.46 \\
			Bi-DiffCD {\scriptsize(IJCAI 25)} \dag  & Unified  & 68.3   & 76.5  & 88.30 & 79.05 & 91.64 & 84.57 & 84.68 & 73.42 & 50.42 & 33.72 & 77.68 & 63.50 & 78.54 \\
			HRSICD {\scriptsize(ISPRS 25)} \dag  & Unified     & 20.5   & 89.3  & 88.05 & 78.66 & 91.29 & 83.98 & 85.23 & 74.26 & 53.58 & 36.60 & 79.51 & 65.99 & 79.53 \\
			HeteCD {\scriptsize(ISPRS 25)} \dag  & Unified     & 50.2   & 58.2  & 89.14 & 80.40 & 91.41 & 84.18 & 86.62 & 76.40 & 53.56 & 36.57 & 78.84 & 65.07 & 79.91 \\
			
			\hline
			\textbf{Ours} \dag     & Unified                         & 52.9   
			& 35.5  & \underline{91.93} & \underline{85.06} & \textbf{95.26} & \textbf{90.94} & \textbf{96.44} & \textbf{93.13} & \textbf{60.43} & \textbf{43.29} & \textbf{81.46} & \textbf{68.71} & \textbf{85.10} \\
			\bottomrule
	\end{tabular}}
	\vspace{-0.2cm}
	\caption{Comparison with state-of-the-art methods. We compare our unified UniRoute against both specialized models (trained individually) and unified baselines (re-implemented under our unified protocol). \dag denotes unified training setting. The best and second-best results are highlighted in \textbf{bold} and \underline{underlined}. All metrics are expressed as percentages (\%).}
	\label{tab:tab2}
\end{table*}

\noindent\textbf{Decision Consistency for Confident Supervision.}
Effective distillation relies on the teacher providing sharp, unambiguous guidance. However, data scarcity often leads to routing ambiguity, where the teacher hesitates between experts. To ensure the teacher generates decisive guidance, we introduce an entropy regularization term on the gating distribution:
\begin{equation}
\mathcal{L}_{ent} = -\frac{1}{|\Omega|}\sum_{u\in\Omega}\sum_{k=1}^{K} p_{u,k}\log(p_{u,k}+\epsilon),
\end{equation}
where $\Omega$ denotes the spatial domain and $K$ is the number of experts. Minimizing this entropy term encourages low-entropy, confident routing decisions,
thereby promoting sparse, near-binary expert selection and structurally distinct distilled knowledge.

\noindent\textbf{Representation \& Statistical Consistency.}
Reliable knowledge transfer further requires stable feature representations. We apply a feature-level consistency loss  that minimizes the cosine distance of deep features in unchanged areas $\Omega_0$:
\begin{equation}
\mathcal{L}_{cons} = \frac{1}{|\Omega_0|} \sum_{u \in \Omega_0}
\left( 1 - \frac{\mathbf{F}_1(u) \cdot \mathbf{F}_2(u)}
{\|\mathbf{F}_1(u)\|_2 \, \|\mathbf{F}_2(u)\|_2} \right),
\end{equation}
where $\mathbf{F}_1(u)$ and $\mathbf{F}_2(u)$ denote the feature vectors at spatial location $u$ from the two input images.
To prevent domain shifts from disrupting this representation alignment, we adopt Domain-Specific Batch Normalization (DSBN), 
which follows standard hard domain routing during training, while enabling adaptive normalization selection at inference.
For each modality $d$, DSBN maintains independent affine parameters $\{\gamma^{(d)}, \beta^{(d)}\}$ and running statistics $\{\mu^{(d)}, \sigma^{(d)}\}$. 
\begin{equation}
\text{DSBN}(x^{(d)}) = \gamma^{(d)} \frac{x^{(d)} - \mu^{(d)}}{\sqrt{(\sigma^{(d)})^2 + \epsilon}} + \beta^{(d)}.
\end{equation}
This isolation prevents the statistical distribution of one modality (e.g., speckle noise in SAR) from contaminating others, creating a stable environment for unified optimization.
The final training objective aggregates these multi-level consistency constraints:
\begin{equation}
\mathcal{L}_{total} = \mathcal{L}_{seg} + \lambda_{cons}\mathcal{L}_{cons} + \lambda_{kd}\mathcal{L}_{kd} + \lambda_{ent}\mathcal{L}_{ent},
\end{equation}
where $\mathcal{L}_{seg}$ is the segmentation loss and $\lambda$ terms balance the auxiliary objectives. In practice, the auxiliary losses are activated in a stage-wise manner.
During unified pre-training, we optimize $\mathcal{L}_{seg}$ together with
$\mathcal{L}_{cons}$ and $\mathcal{L}_{ent}$ to stabilize expert routing.
During the CASD fine-tuning stage, $\mathcal{L}_{ent}$ is removed and
$\mathcal{L}_{kd}$ is introduced to perform self-distillation under geometric transformations.

\section{Experiments and Analysis}

\subsection{Datasets and Implementation Details}
We evaluate our UniRoute on five public datasets: LEVIR-CD~\cite{LEVIR} and WHU-CD~\cite{WHU} (Optical-Optical); HTCD~\cite{SUNet} (Optical-UAV); and MT-Wuhan~\cite{MT-WHU} and XiongAn~\cite{XiongAn} (heterogeneous Optical-SAR). 
\begin{figure*}[t!]
	\centering
	\includegraphics[width=1\linewidth]{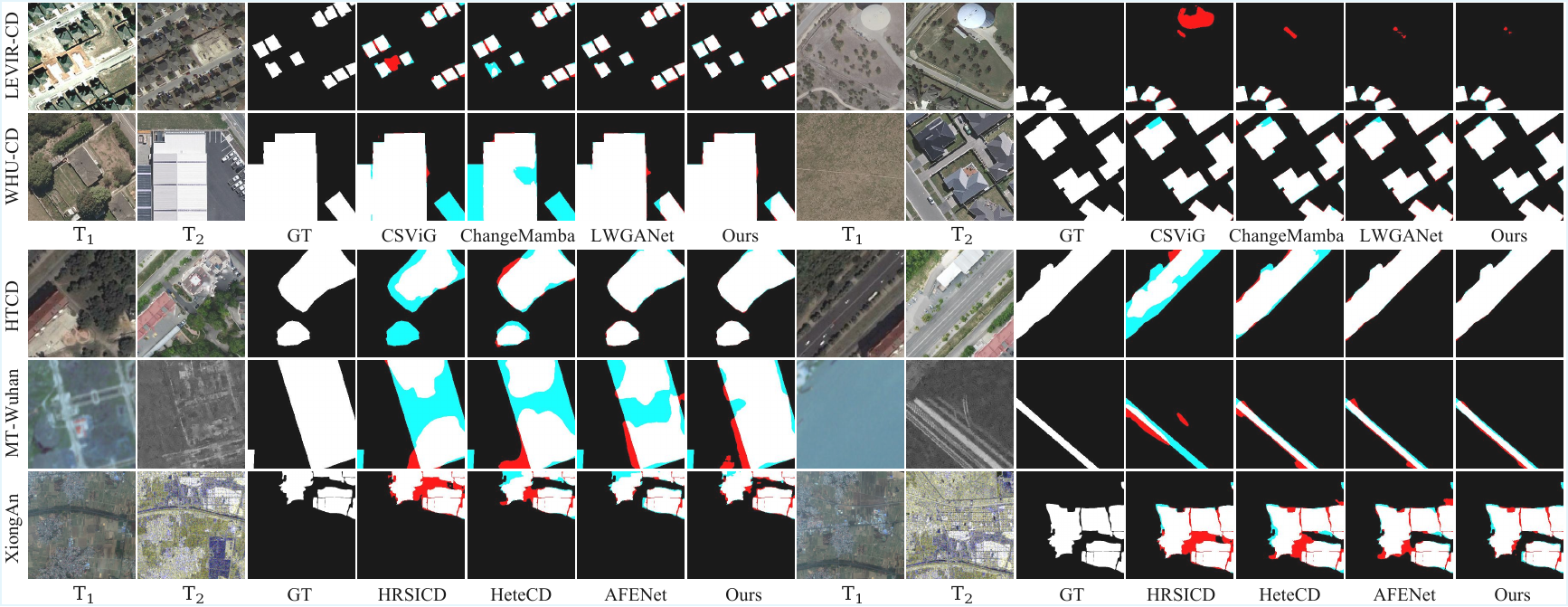} 
	\vspace{-0.6cm}
	\caption{ Qualitative comparisons on five datasets.
		We show predicted change maps of representative methods and our UniRoute on both homogeneous and heterogeneous settings.
	}
	\vspace{-0.2cm}
	\label{fig:compare_SoTA}
\end{figure*}
We use a ResNet50~\cite{ResNet50} backbone pretrained on ImageNet. Training follows a two-stage paradigm: (1) \textit{Unified Pre-training} for 200 epochs on combined datasets using AdamW (learning rate = $3 \times 10^{-4}$); (2) \textit{CASD Fine-tuning} for 15 epochs with reduced lr ($1 \times 10^{-5}$) 
and re-balanced sampling for small-sample datasets. We report F1-score and IoU on the test sets using the best validation checkpoint. 
For fair comparison, all baseline methods are re-implemented using their official codebases and standard protocols. 
Note that we do not apply CASD to baselines.
This is because CASD is tailored to regularize dynamic routing instability; applying strict geometric consistency to static backbones often leads to optimization conflicts or marginal gains in our preliminary tests.
 More  details are  in the  Supplementary Material.

\subsection{Comparison with State-of-the-Art Methods}

We compare UniRoute with recent SOTA methods, including CSViG~\cite{CSViG}, ChangeMamba~\cite{ChangeMamba}, LGCANet~\cite{LGCANet}, LWGANet~\cite{LWGANet}, AFENet~\cite{AFENet}, Bi-DiffCD~\cite{Bi-DiffCD}, HRSICD~\cite{HRSICD} and HeteCD~\cite{XiongAn}. Tab.~\ref{tab:tab2} shows the quantitative results. More details are in the Supplementary Material.

\noindent\textbf{Comparison with Specialized Models.} 
Specialized models perform well in narrow domains, but they lack versatility.
For example, LWGANet achieves a high F1 score of 92.42\% on LEVIR-CD.
However, the same model drops sharply to 55.78\% on MT-Wuhan.
To reach the coverage of UniRoute, one would need an ensemble of best specialists.
This ensemble requires 132.1M parameters and 332.7G FLOPs
while our UniRoute breaks this trade-off.
With only 52.9M parameters, UniRoute matches the ensemble in average performance.
It achieves an Avg F1 of 85.10\%, compared to 84.76\% for the ensemble.
As a result, our model achieves SOTA accuracy with only 40\% of the parameters and 10\% of the FLOPs of the ensemble.

\noindent\textbf{Comparison with Unified Baselines.} We observe that retraining  SOTA models on mixed datasets frequently leads to performance degradation due to negative transfer.  ChangeMamba drops from 91.18\% to 89.35\% on LEVIR-CD  and from 56.44\% to 53.88\% on MT-Wuhan in the unified setting.  In contrast, UniRoute remains robust, which outperforms the best unified baseline (AFENet) by 4.64\% on average. The gains on MT-Wuhan (+4.32\%) and HTCD (+7.82\%) highlight the efficacy of dynamic routing in resolving conflicts. 

\noindent\textbf{Qualitative Analysis.}
As shown in Fig.~\ref{fig:compare_SoTA},
On homogeneous LEVIR-CD and WHU-CD datasets, ChangeMamba and CSViG often exhibit blurred boundaries and noticeable prediction errors.
For heterogeneous scenarios (e.g. MT-Wuhan), AFENet misclassifies speckle noise as changes. 
In contrast, UniRoute delivers clean shapes and preserves fine-grained edges.
Notably, in HTCD, our model maintains robust alignment despite viewpoint distortions, avoiding false alarms common in subtraction-based approaches.
\subsection{Ablation Study}

As shown in Tab.~\ref{tab:abstudy}, we conduct ablation studies on the representative datasets LEVIR-CD, HTCD, and MT-Wuhan to analyze model behavior under different settings. More details are in the Supplementary Material.

\noindent\textbf{Effectiveness of AR$^2$-MoE.} 
To demonstrate the effectiveness of AR$^2$-MoE, we first remove AR$^2$-MoE from the network.
Tab.~\ref{tab:abstudy} reports the corresponding results.
This design causes a slight performance drop on LEVIR-CD and WHU-CD, but leads to a  degradation on MT-Wuhan, where the F1  decreases to 55.8\%.
As shown in Row \#2-\#4, introducing AR$^2$-MoE leads to consistent performance gains. 
We tend to insert AR-MoE at different stages of the network.
The experiments show that the accuracy consistently improves as AR-MoE is introduced.
Considering the trade-off between computational cost and accuracy, we place AR-MoE only at the deeper Stages 2–4.
Rows \#05 and \#06 further show that relying on a single expert alone,
either local-detail or global-context,
is insufficient to handle heterogeneous scenes,
highlighting the necessity of dynamic expert selection.
Multi-kernel fusion (Row \#07) improve capacity by combining multiple receptive fields, but they indiscriminately apply the same fusion. By contrast, AR$^2$-MoE uses conditional routing to adapt the receptive field on a per-pixel basis, allowing different regions to rely on receptive fields of appropriate scales.
\begin{table}[t]
	\centering

	\resizebox{\columnwidth}{!}{
		\begin{tabular}{llccc}
			\toprule
			NO.  & Variants                       & LEVIR-CD             & HTCD                 & MT-Wuhan             \\
			\midrule
			\#00 & \textbf{UniRoute (Ours)}       & \textbf{91.93}       & \textbf{96.44}       & \textbf{60.43}       \\
			\hline
			\rowcolor{gray!20}
			\multicolumn{5}{l}{\textit{(a) AR$^2$-MoE}}                                  \\
			\#01 & w/o AR$^2$-MoE (Vanilla ResNet)  & 90.34                & 95.30                & 55.80                \\
			\#02 & w/ AR$^2$-MoE (Stage 4 only)     & 91.45                & 95.65                & 57.20                \\
			\#03 & w/ AR$^2$-MoE (Stage 3, 4)       & 91.68                & 96.10                & 58.90                \\
			\#04 & w/ AR$^2$-MoE (Stage 1, 2, 3, 4) & 91.86                & 96.40                & 59.79                \\
			\#05 & w/ Only \textit{Local-Detail Expert}  & 91.45                & 95.80                & 56.20                \\
			\#06 & w/ Only \textit{Global-Context Expert}       & 90.22                & 95.10                & 59.80                \\
			\#07 & w/  Multi-Kernel (3×3 + 5×5 + 7×7)  & 90.70 & 95.60 &58.45 \\
			\hline
			\rowcolor{gray!20}
			\multicolumn{5}{l}{\textit{(c) MDR-MoE}} \\
			\#08 & w/o MDR-MoE (Naive Concat) & 90.50 & 95.80 & 56.50 \\
			\#09 & w/ Sub. Only (Static) & 91.55 & 90.50 & 42.10 \\
			\#10 & w/ Cat. Only (Static) & 90.40 & 95.63 & 58.50 \\
			\#11 & w/ Mul. Only (Static) & 89.20 & 91.23 & 54.00 \\
			\#12 & w/ DFC (Sub+Cat+Mul) & 91.60 & 94.87 & 57.80 \\
			\hline
			\rowcolor{gray!20}
			\multicolumn{5}{l}{\textit{(d) CASD}} \\
			\#13 & w/o CASD (Standard Training) & 91.80 & 96.20 & 57.78 \\
			\hline
			\rowcolor{gray!20}
			\multicolumn{5}{l}{\textit{(e) Different Backbones}} \\
			\#14 & ResNet18 & 90.10 & 94.50 & 54.20 \\
			\#15 & ResNet34 & 91.20 & 95.80 & 58.10 \\
			\#16 & ResNet50 (Ours) & 91.93 & 96.44 & 60.43 \\
			\#17 & Swin-T & 91.60 & 96.00 & 57.50 \\ \bottomrule
	\end{tabular}}
	\vspace{-0.2cm}
	\caption{Ablation study on key components and architectures. 
	}
	\vspace{-0.1cm}
	\label{tab:abstudy}
\end{table}
\begin{figure}[t]
	\centering
	\includegraphics[width=1\linewidth]{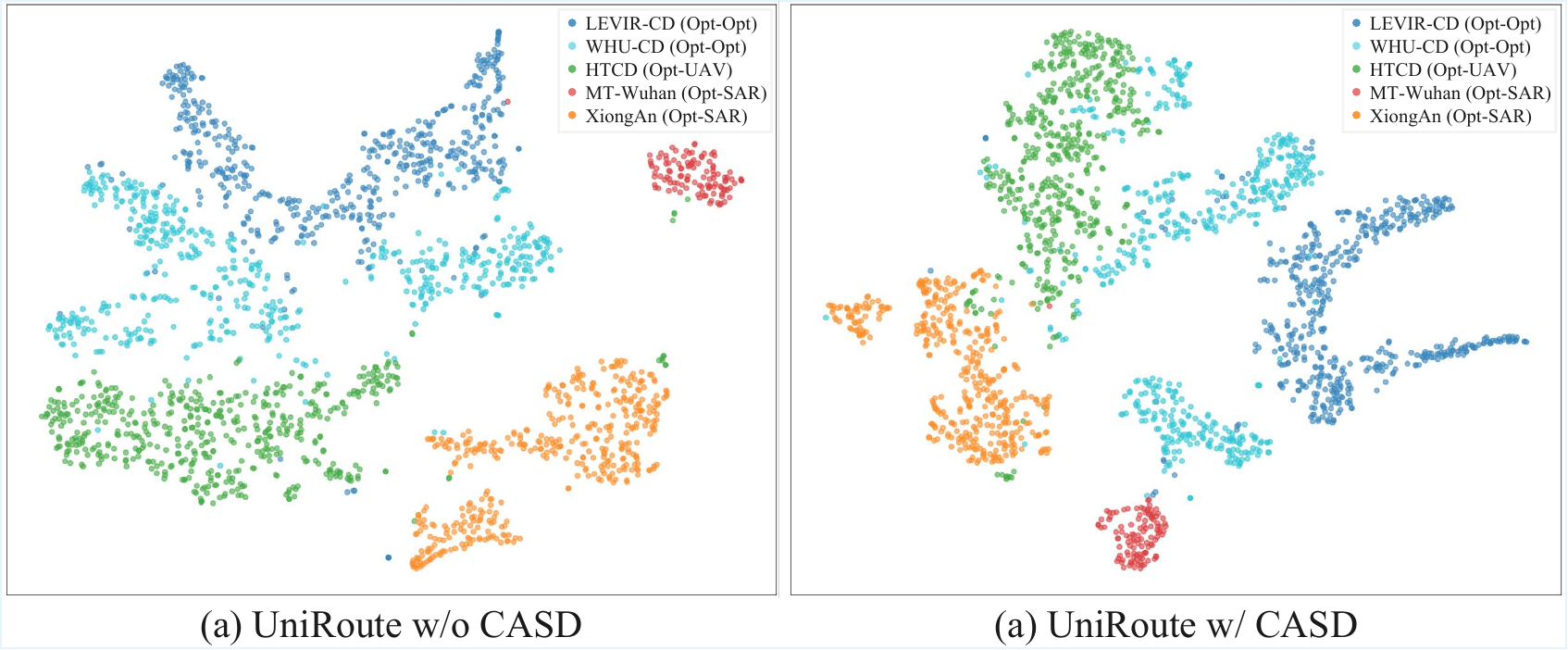} 
	\vspace{-0.6cm}
	\caption{t-SNE visualization of feature distributions. 
	}
	\vspace{-0.2cm}
	\label{fig:tsen}
\end{figure}
\begin{table}[t]
	\centering
	
	\resizebox{\columnwidth}{!}{
		\begin{tabular}{l|c|ccc}
			\toprule
			Routing Mechanism & Differentiable & LEVIR-CD & HTCD & MT-Wuhan  \\
			\midrule
			Soft Gating (Weighted Sum) &$\checkmark$  & 91.50 & 96.15 & 58.50  \\
			Top-1 Hard (No STE) &$\times$  & 88.40 & 92.10 & 52.10  \\
			Gumbel-Softmax &$\checkmark$  & 91.65 & 96.22 & 59.20  \\
			\textbf{STE Hard Routing (Ours)} &$\checkmark$  & \textbf{91.93} & \textbf{96.44} & \textbf{60.43}  \\
			\bottomrule
		\end{tabular}
		
	}
	\vspace{-0.2cm}
	\caption{Analysis of routing strategies. 
	}
	\label{tab:routing_strategy}
\end{table}
\begin{table}[t]
	\centering
	
	\resizebox{\columnwidth}{!}{
		\begin{tabular}{l|c|c|cc}
			\toprule
			Initialization & Data Fraction & Epochs & F1  & $\Delta$ \\
			\midrule
			ImageNet Init. & 5\% ($\sim$800 pairs) & 50 & 69.25 & - \\
			\textbf{UniRoute Init. (Ours)} & \textbf{5\%} & \textbf{20} & \textbf{74.88} & \textbf{+5.63} \\
			\midrule
			ImageNet Init. & 10\% ($\sim$1600 pairs) & 50 & 74.97 & - \\
			\textbf{UniRoute Init. (Ours)} & \textbf{10\%} & \textbf{20} & \textbf{77.48} & \textbf{+3.16} \\
			\bottomrule
		\end{tabular}
	}
	\vspace{-0.2cm}
	\caption{Few-Shot transfer learning on unseen SYSU-CD dataset. 
	}
	\vspace{-0.2cm}
	\label{tab:transfer_learning}
\end{table}

\noindent\textbf{Effectiveness of MDR-MoE.}
To demonstrate the effectiveness of MDR-MoE, we first remove MDR-MoE from the network.
The performance drops on all three datasets.
As shown in Row \#9-\#11, different modalities favor different difference operators.
The subtraction operator performs well on the homogeneous LEVIR-CD dataset.
However, its performance on MT-Wuhan drops sharply to 42.1\%.
This result indicates that SAR features introduce strong noise during differencing.
To compare MDR-MoE with existing difference operators, we replace our design with the DFC module from ChangeCLIP.
DFC integrates multiple operators into a unified difference module.
Row \#12 shows that DFC achieves competitive performance.
However, our method still outperforms DFC by 1.57\% on HTCD and 2.63\% on MT-Wuhan. This shows that the limitation lies not in the primitives themselves but in the fusion strategy: statically including noisy branches contaminates features, whereas MDR-MoE effectively prunes them.

\noindent\textbf{Effectiveness of CASD.} 
As shown in Row \#13, we first remove the CASD training strategy.
Tab.~\ref{tab:abstudy} shows that the performance on LEVIR-CD and HT-CD drops only slightly. However, the performance on MT-Wuhan decreases to 57.78\%. This result further confirms that CASD prevents overfitting on small-scale heterogeneous datasets while provides positive gains on other datasets.    It is noted UniRoute \textit{without} CASD still outperforms the strongest unified baseline (AFENet) on challenging heterogeneous datasets (e.g., 57.78\% vs. 56.11\% on MT-Wuhan), confirming that our architectural advantage is intrinsic.
The t-SNE visualization in Fig.~\ref{fig:tsen} supports this conclusion.
CASD training strategy effectively reduces modality discrepancies.
The embeddings of homogeneous and heterogeneous datasets become closer in the feature space.


\noindent\textbf{Backbone Scalability.}
We replace the backbone of UniRoute with different architectures, including ResNet variants and transformer.
Rows \#14–\#17 in the table report the corresponding results.
ResNet-50 and Swin-T achieve the best overall performance.
Considering the trade-off between accuracy, parameter count, and computational cost, we adopt ResNet-50 as the backbone in this work.

\noindent\textbf{ Routing Strategies Analysis.}
To evaluate the impact of different gating strategies, we replace STE with alternative designs.
The results in Tab.~\ref{tab:routing_strategy} show that standard soft gating achieves reasonable performance.
However, this strategy performs poorly on heterogeneous data such as Optical–SAR.
Top-1 selection without STE fails to converge due to gradient blocking
and yields low accuracy.
By introducing STE, our method enables end-to-end training with discrete selection. This design balances sparsity and differentiability, which achieves the best performance.
\subsection{Generalization and Data Efficiency}

Due to severe domain shift and modality-specific noise, fully zero-shot CD across  modalities is highly challenging.
Therefore, we conduct few-shot transfer experiments in this work.
We fine-tune the model on the previously unseen SYSU dataset using limited training data.
The experimental results in Tab.~\ref{tab:transfer_learning} show that our method achieves a 5.63\% improvement over ImageNet initialization when using only 5\% of the data.
These results indicate that our model learns change representations with strong cross-dataset transferability.

\section{Conclusion}
In this paper, we identify the intrinsic conflicts in receptive fields and fusion operations that hinder unified CD. To resolve them, we propose UniRoute, a modality-adaptive framework that formulates feature extraction and fusion as a conditional routing problem. Through AR$^2$-MoE and MDR-MoE,  our model adaptively selects receptive fields and  optimal fusion primitives to accommodate heterogeneous characteristics at the pixel level. 
Furthermore, 
we introduce a CASD strategy to regularize training
under data-scarce and heterogeneous settings by enforcing multi-level consistency.
Extensive experiments show that UniRoute consistently outperforms unified baselines and remains competitive with strong specialists,
while keeping a single unified model that is efficient in both parameters and computation.



\appendix

\bibliographystyle{named}
\bibliography{ijcai26}

\end{document}